\documentclass{tETA2e}
\usepackage{graphicx}
\usepackage{amsmath}
\usepackage{amssymb}
\usepackage{algorithmic}
\usepackage{algorithm}
\usepackage{array}
\usepackage{multirow,tabularx}
\usepackage{epstopdf}
\usepackage{subfigure}
\usepackage{url}
\usepackage{color}
\usepackage[pagewise]{lineno}

\usepackage[longnamesfirst,sort]{natbib}
\bibpunct[, ]{(}{)}{;}{a}{,}{,}
\renewcommand\bibfont{\fontsize{10}{12}\selectfont}

\theoremstyle{plain}

\theoremstyle{definition}

\theoremstyle{remark}

\newcommand{\bfa}{{\textbf{a}}}

\newcommand{\bfomega}{{\boldsymbol{\omega}}}
\newcommand{\bfx}{{\textbf{x}}}

\newcommand{\bfw}{{\textbf{w}}}
\newcommand{\bfy}{{\textbf{y}}}

\newcommand{\bfz}{{\textbf{z}}}

\newcommand{\bff}{{\textbf{f}}}

\newcommand{\bfpi}{{\boldsymbol{\pi}}}

\newcommand{\bfupsilon}{{\boldsymbol{\upsilon}}}

\begin{document}



\title{Domain transfer convolutional attribute embedding}

\author{
\name{Fang Su\textsuperscript{a}$^{\ast}$
\thanks{$^\ast$Corresponding author.},
Jing-Yan Wang\textsuperscript{b}}
\affil{
\textsuperscript{a}
School of Economics and Management, Shaanxi University of Science \& Technology, Xi'an, ShaanXi Province, P.R.C, 710021\\
\textsuperscript{b}
New York University Abu Dhabi, Abu Dhabi, United Arab Emirates}
\received{v5.0 released July 2015}
}

\maketitle

\begin{abstract}
In this paper, we study the problem of transfer learning with the attribute data. In the transfer learning problem, we want to leverage the data of the auxiliary and the target domains to build an effective model for the classification problem in the target domain. Meanwhile, the attributes are naturally stable cross different domains. This strongly motives us to learn effective domain transfer attribute representations. To this end, we proposed to embed the attributes of the data to a common space by using the powerful convolutional neural network (CNN) model. The convolutional representations of the data points are mapped to the corresponding attributes so that they can be effective embedding of the attributes. We also represent the data of different domains by a domain-independent CNN, ant a domain-specific CNN, and combine their outputs with the attribute embedding to build the classification model. An joint learning framework is constructed to minimize the classification errors, the attribute mapping error, the mismatching of the domain-independent representations cross different domains, and to encourage the the neighborhood smoothness of representations in the target domain. The minimization problem is solved by an iterative algorithm based on gradient descent. Experiments over benchmark data sets of person re-identification, bankruptcy prediction, and spam email detection, show the effectiveness of the proposed method.
\end{abstract}


\section{Introduction}

\subsection{Backgrounds}

In the machine learning problems, domain transfer learning has recently attracted much attention \citep{DBLP,wang2017unleash,Deep2018124,Wang2017,Yang2017,Zhang2017242}. Transfer learning refers to the learning problem of a predictive model for a target domain, by leveraging the data from both the target domain and one or more auxiliary domains. The target domain is in lack of class labels, which makes the learning in the target domain difficult. The auxiliary domains have the same input space and the label space, however, the data distribution of the auxiliary domains are significantly different from the target domain, thus the auxiliary domain data cannot be directly used to learn the model in the target domain. To solve this problem, domain-transfer learning is proposed to transfer the data representation and/or the models of the auxiliary domains to fit the data of the target domain, so that the classification performance of the target domain can be improved. For example, in the problem of person re-identification, to identify one person captured by one camera, we learn a classifier to predict the ID of the image of a person. Usually, there are more than one cameras, and we can use the data of different cameras to help the learning for one target camera. Because of the angle of different cameras are different, the data of the multiple cameras cannot be directly combined. Thus the transfer learning technology is needed to leverage the gaps between the cameras \citep{IbnKhedher201794,An201739,Zhao2017218,Hassen201811}.

One shortage of traditional transfer learning methods is the that the attributes of the data are not used by the classification model. But the attributes of the data actually has the nature of stability across the domains. For example, in the problem of person re-identification, because of the change of the angles of the cameras, the appearances of the same person in different cameras may change, the attributes usually keep stable, such as the attribute of long hair, wearing short pans, and/or carrying a bag. Thus using the attributes of the data is critical for the transfer learning \citep{Kulkarni2014220,Suzuki201475,Suzuki20143627,Peng2017}. In this paper, we study the problem of effective use of both input data and attribute for the domain-transfer learning problem, and propose a novel method of attribute embedding based on the popular convolutional neural network (CNN) \citep{shen2017multi,shen2015multi,geng2017novel,zhang2017learning,geng2016learning,Jing20171,Puri2018326,RoaBarco2018377,Todoroki2018140,Waijanya2018179,Fujino2018278} to solve this problem. Further, we develop a novel model using the attribute embedding as the input for the learning of the target domain classification model.

\subsection{Relevant works}

Our work is an effective representation method of attributes for the problem of transfer learning problem. However, there only two existing works in this direction, and we introduce them as follows.

\begin{itemize}
\item Peng et al. \citep{Peng2017} proposed to represent the attribute vectors of each data point by using an attribute dictionary. Each data point is reconstructed by the elements in the dictionary, and the reconstruction coefficients are used as the new representation of the attributes. The attribute vector of a data point is mapped to the new representation vector by a linear transformation matrix so that the new representation vector is linked to the attribute vector. To leverage the auxiliary and the target domains, the same attribute representation method is applied to both auxiliary and target domains. The learning process is regularized by the class-intra similarity in the auxiliary domains, and by the neighborhood in the target domain.

\item Su et al. \citep{Su2017} proposed a low-rank attribute embedding method for the problem of person re-identification of multiple cameras. The proposed method tries to solve the problem of multiple cameras based person re-identification as a multi-task learning problem. The proposed method uses both the low-level features with mid-level attributes as the input of the identification model. The embedding of attributes maps the attributes to a continuous space to explore the correlative relationship between each pair of attributes and also recovers the missing attributes.
\end{itemize}

Both these two methods of attribute representation are based on the linear transformation. However, a simple linear function may be insufficient to represent the attributes effectively. In the domain transfer area, a group of methods \citep{meng_ICIP2012,meng_FG2013,meng_cybernetics2015,meng_Xin_2016} embedded a physical structure of high dimensional data into another domain with low dimensionality using a non-linear mapping, which is trained by balancing the effect of data and a heuristic physical prior. These methods inspired us to embed the attributes of the data to a common space.

\subsection{Our contribution}

In this paper, we propose a novel attribute embedding method for attributes for the problem of domain transfer learning. The embedding of attributes is based on CNN model. The convolutional output of the input data is further mapped to the attribute vector. In this way, the attribute embedding vector not only represents the attributes of a data point but also contains the pattern of the input data constructed by the CNN model, which has been proven to be a powerful representation model. To construct the classification model for each domain, we also learn a domain-independent convolutional representation and a domain-specific convolutional representation. The domain-independent convolutional representation maps the data of different domains to a shared data space to capture the patterns shared over all the domain. The domain-specific convolutional representation is used to represent the patterns specifically contained by each domain. The classification model of each domain is based on the three types of convolutional representations, i.e., attribute embedding, domain-independent and domain-specific representations. To learn the parameters of the models, we propose to minimize the mapping errors of the attributes, the classification errors across different domains, the mismatching of different domains in the domain-independent representation space, and the dissimilarity between the neighboring data points in the target domain. The joint minimization problem is solved by an alternate optimization strategy and the gradient descent algorithm.

\subsection{Paper organization}

This paper is organized as follows. In section \ref{sec:method}, we introduce the proposed model and the learning method of the parameters of the model. In section \ref{sec:exp}, we test the proposed method over some benchmark datasets, compare it to the state-of-the-art methods. In section \ref{sec:conclusion}, we give the conclusion of this paper.

\section{Method}
\label{sec:method}

In this section, we will introduce the proposed method of attribute embedding and cross-domain learning.  The proposed model and the corresponding learning problem is firstly introduced, and then the optimization method is developed to solve the learning problem. Finally, we give an iterative algorithm based on the optimization results.

\subsection{Problem embedding}

In the problem setting of cross-domain learning, we assume we have $T$ domains. The first $T-1$ domain are the auxiliary domains, while the $T$-th domain is the target domain. The problem is to learn an effective model for the classification of target domain. The input data for the training is given as follows.
\begin{itemize}
\item The input data sets of the $T$ domains are denoted as $\mathcal{X}^t|_{t=1}^{T}$, where $\mathcal{X}^t = \{X_i^t\}$ is the data set of the $t$-th domain, and $X_i^t = \left[\bfx_{i1}^t,\cdots, \bfx_{i|X_i^t|}^t\right]\in R^{d\times |X_i^t|}$ is the input matrix of the $i$-th data point of the $t$-th domain, and each column of the matrix is a feature vector of a instance.
\item Moreover, for each data point $X_i^t$, an attribute vector is attached, $\bfa_i^t = \left[a_{i1}^t, \cdots, a_{i|a|}^t \right] \in \{1,0\}^{|a|}$, where $a_{ij}^t = 1$ if the $i$-th data point of the $t$-th domain has the $j$-th attribute, and $0$ otherwise.
\item Meanwhile, all the data points of the auxiliary domains, and a small part of the data points of the target domain has the class label vector. For a data point, $X_i^t$, a class label vector $\bfy_i^t = \left[\bfy_{i1}^t, \cdots, \bfy_{i|y|}^t\right] \in \{1,0\}^{|y|}$, where $\bfy_{ij}^t = 1$ if $X_i^t$ belongs to the $j$-th class, and $0$ otherwise.
\end{itemize}

Our model is composed of three convolutional representation sub-models, namely the convolutional attribute embedding model, the domain-independent representation model, and the domain-specific convolutional representation model. The convolutional attribute embedding model is used to leverage the input instances and the attributes, thus its learning is regularized by the attributes of the given data points. The domain-independent representation is a model for the data of all domains, and its main function is to map the data of different domains to a common space, thus it is regularized by the domain-matching term. The domain-specific convolutional representation is designed for different domains to handle the discrepancy of data of different domains. The representation of an input data point is the combination of the outputs of the three models, and it is used to predict the class label, and meanwhile, it is also regularized the neighborhood in the target domain. Accordingly, to construct the classification model, we consider the following problems.

\textbf{Convolutional attribute embedding} We propose to embed the attributes of each input data point to a vector, and use the convolutional representation of the input data as the embedding vector. The embedding vectors will be further used as input of the classification model. Given the the input matrix of a data point, $X = [\bfx_1,\cdots, \bfx_{|X|}]$, to obtain its convolutional representation, we have a four-step process:

\begin{enumerate}
\item We first use a sliding window of $\alpha$ instances, and concatenate the instances within the window to a new vector $\bfz \in R^{\alpha d}$. With the sliding window moving from the beginning to the end with a step of 1 instance, we obtain the new input data matrix,

\begin{equation}
\begin{aligned}
Z = [\bfz_1, \cdots, \bfz_{|X|+\alpha-1}],
\end{aligned}
\end{equation}
where $\bfz_i = \begin{bmatrix}
\bfx_i\\
\vdots \\
\bfx_{i+\alpha-1}
\end{bmatrix} \in R^{\alpha d}$ is the output of the $i$-th step of sliding. It is the concatenation of vectors of the $i$-th instance to the $i+\alpha-1$-th instance.

\item Then a filter matrix $W_a \in R^{(\alpha d)\times m }$ is applied to the new input matrix, where each column of the filter matrix is a filter vector. The filtering result is the multiplication between $W_a^\top$ and $Z$, $W_a^\top Z$.

\item Filtering is followed by an activation operation. In the activation operation, each element of the input matrix is transformed by a non-linear activation function, which is defined as the Rectified Linear Units (ReLU) $g(x) = \max(0,x)$. The output of activation operation is denoted as $g(W_a^\top Z)$.

\item The last step is max-pooling. Given the output of the activation operation, we select the maximum element from each row, the output is denoted as $\max \left( g(W_a^\top Z) \right)$, where $\max(X)$ is a row-wise maximization operator.

\end{enumerate}
The overall output of the convolutional representation can be obtained by the chain function of the four steps, denoted as

\begin{equation}
\label{equ:f_a}
\begin{aligned}
f_a(X)
&= \max \left( g(W_a^\top Z) \right)\\
&=
\begin{bmatrix}
\max(g(\bfw_{a1}^\top Z)) \\
\vdots \\
\max(g(\bfw_{am}^\top Z))
\end{bmatrix}
\end{aligned}
\end{equation}
Since this convolutional representation of $X$ is used as its attribute vector embedding, we propose to link it to the attribute vector $\bfa$ by a linear mapping function,

\begin{equation}
\begin{aligned}
f_a(X) \leftarrow \Theta ^\top \bfa,
\end{aligned}
\end{equation}
where $\Theta \in R^{|a|\times m}$ is the mapping matrix. The convolutional representation of $X$ is generated from its original data instances, meanwhile it is a mapping of the attributes of the input data. In this way, the embedding of the attributes only has the property of the attribute properties themselves, but also relies on the effective convolutional representation of the input data itself. Thus the embedding leverage the convolutional representation and the attributes well.

To reduce the mapping errors, we proposed to minimize the Frobenius norm distance between the convolutional representations and the mapping results for all the data points of all domains,

\begin{equation}
\begin{aligned}
\min_{\Theta, W_a}  \sum_{t=1}^{T} \left ( \sum_{i=1}^{n_t} \left \|f_a(X_i^t) - {\Theta} \bfa_i^t\right \|_F^2 \right ),
\end{aligned}
\end{equation}
{where $\Theta$ is the mapping matrix of the linear mapping function for attribute vectors.} By minimizing this objective, we obtain an effective embedding of the attributes of the input data. Please note that the embedding is for the attribute vector, instead for each of the attribute element.

\textbf{Domain-independent representation} To predict the class labels for the data points in multiple domains, we proposed the data of each domain to represent the data into a base convolutional representation, and a domain-specific convolutional representing. The base convolutional representation function is shared across all the domains. It tries to extract features relevant to the class labels, but independent of the specific domain. The base convolutional recreation function is also based on the sliding window, filtering, activation, and max-pooling. The base convolutional representation of $X$ is defined as,

\begin{equation}
\begin{aligned}
f_0(X) = \max \left( g(W_0^\top Z) \right),
\end{aligned}
\end{equation}
where $Z$ is the output of sliding window, and $W_0$ is the filter matrix of the base convolutional representation function.

Since the base convolutional representation is domain-independent, we hope the representations of data points from different domains can be similarity to each other. To this end, we impose that the distribution of the base representations of different domains is of the same. We use the mean vector of the representations of each domain as the presentation of the distribution of the domain. For the $t$-th domain, the mean vector is given as $\frac{1}{n_t}\sum_{i=1}^{n_t} f_0(X_i^t)$. To reduce the mismatch among the domains, we proposed to minimize the Frobenius norm distances between the mean vectors of each pair of domains.

\begin{equation}
\begin{aligned}
\min_{W_0}
\sum_{t,t'=1, t<t'}^{T} \left\| \frac{1}{n_t}\sum_{i=1}^{n_t} f_0(X_i^t)
- \frac{1}{n_{t'}}\sum_{i=1}^{n_{t'}} f_0(X_i^{t'})  \right \|_F^2.
\end{aligned}
\end{equation}
By minimizing this problem, we hope the base convolutional representation function can map data points of different domain to a common shared data space.

\textbf{Cross-domain class label estimation}
To predict the class labels for the data points of different domains, we also consider the representation of the data points according to the domains. This is the domain-specific representations. The representation is also based on convolutional network function, and the function of the $t$-th domain of a data point $X$ is given as,

\begin{equation}
\label{equ:f}
\begin{aligned}
f_t(X) = \max \left( g(W_t^\top Z) \right),
\end{aligned}
\end{equation}
where $W_t$ is the filter matrix of the $t$-th domain specific convolutional representations.

To estimate the class label from a data point of the $t$-th domain, we combine both the domain-independent and domain-specific convolutional representations of the input data, $f_0(X)$ and $f_t(X) $, and also the attribute embedding of the data, $f_a(X)$. They are concatenated to a longer vector,

\begin{equation}
\label{equ:f}
\begin{aligned}
f(X) = \begin{bmatrix}
f_0(X) \\
f_t(X) \\
f_a(X)
\end{bmatrix} \in R^{m_0+m_t+m_a},
\end{aligned}
\end{equation}
and the longer vector is transformed to a $|y|$-dimensional vector of scores of classification by a matrix $U = \begin{bmatrix}
U_0 \\
U_t \\
U_a
\end{bmatrix} \in R^{(m_0+m_t+m_a)\times |y|}$ in a classification function,

\begin{equation}
\label{equ:h}
\begin{aligned}
h_t(X) = U^\top f(X) = U_0^\top f_0(X) + U_t^\top f_t(X) + U_a^\top f_a(X) , t=1,\cdots,T,
\end{aligned}
\end{equation}
where $U_0$, $U_t$, and $U_a$ are the transformation matrices for the domain-independent representation, domain-specific representation, and the attribute embedding. The classification function is used to predict the class labels, thus we propose to reduce the prediction errors measured by the Frobenius norm distance between the class label vectors and the outputs of $h_t(X)$ for the data pints with available label vectors,

\begin{equation}
\begin{aligned}
\min_{U_t,W_t|_{t=0}^{T}, U_a, W_a}
\left \{
\sum_{t=1}^{T-1}\left(\sum_{i=1}^{n_t}  \|\bfy_i^t - h_t(X_i^t)\|_F^2\right)
+\sum_{i=1}^{l_T}  \|\bfy_i^T - h_t(X_i^T)\|_F^2
\right \}.
\end{aligned}
\end{equation}
Please in the objective function of this problem, for the auxiliary domains, all the data points are labeled, but for the target domain, only the first $l_T$ data points are labeled. Thus for the target domain, we only consider the first $l_T$ data points.

\textbf{Neighbourhood similarity regularization}
For the unlabeled data points in the target domain, we also regularize them by imposing their representations to be constant with the labeled data points in the neighborhood, so that the supervision information can also be propagated to them. To this end, we hope for any neighboring two data points in the target domain, their overall representation vectors are close to each other. We propose to minimize the Frobenius norm distance between the representations of neighboring data points in the target domain,

\begin{equation}
\begin{aligned}
\min_{U_T,W_T, U_a, W_a}
\sum_{i,i'=1}^{n_T} M_{ii'} \left \|f(X_i^T) - f(X_{i'}^T)\right \|_F^2,
\end{aligned}
\end{equation}
where $M_{ii'} =  1$ if $X_i$ and $X_{i'}$ are neighbor to each other, and $0$ otherwise. In this way, if a data point is not labeled, but its representation is also regularized by the other representations of the target domain, especially the representations of the labeled data points. Thus the learning of the unlabeled data points is also benefiting from the labels.

\subsection{Problem optimization}

The learning framework is constructed by combining the learning problems mentioned above,

\begin{equation}
\label{equ:obj1}
\begin{aligned}
\min_{U_t,W_t|_{t=0}^{T}, U_a, W_a, \Theta}
&
\left \{o(U_t,W_t|_{t=0}^{T}, U_a, W_a, \Theta) =
\sum_{t=1}^{T-1}\left(\sum_{i=1}^{n_t}  \|\bfy_i^t - h_t(X_i^t)\|_F^2\right)
\right .\\
&
+\sum_{i=1}^{l_T}  \|\bfy_i^T - h_t(X_i^T)\|_F^2\\
&
+ C_1 \sum_{t=1}^{T} \left ( \sum_{i=1}^{n_t} \left \|f_a(X_i^t) - \Theta \bfa_i^t\right \|_F^2 \right )\\
&
+ C_2 \sum_{t,t'=1, t<t'}^{T} \left\| \frac{1}{n_t}\sum_{i=1}^{n_t} f_0(X_i^t)
- \frac{1}{n_{t'}}\sum_{i=1}^{n_{t'}} f_0(X_i^{t'})  \right \|_F^2\\
&\left.
+ C_3 \sum_{i,i'=1}^{n_T} M_{ii'} \left \|f(X_i^T) - f(X_{i'}^T)\right \|_F^2 \right\},
\end{aligned}
\end{equation}
where $o$ is the objective function, and $C_k, k=1,\cdots,3$ are the tradeoff weights of different regularization terms. This joint learning framework can learn the effective representations of the input data of different domains. The three types of representations are all based on convolutional networks. The attribute embedding are regularized the attribute vectors. The domain-independent representations are regularize by both the labels and the mismatching of the distributions of different domains. The domain-specific representations are only regularized by the labels of the corresponding domains. For the target domain, all the representations are regularized by the neighborhood structure.

To solve this problem, we proposed to use the alternate optimization method. When one parameter is being optimized, others are fixed. In an iterative algorithm, the parameters are updated to optimize the problem alternately. In the following sections, we will discuss how to update the parameters respectively.

\subsubsection{Optimization of filters of $f_a(X)$}

When the filters of $W_a$ is optimized, we substitute (\ref{equ:f_a}), (\ref{equ:f}) and (\ref{equ:h}) to (\ref{equ:obj1}), and remove all the terms which are irrelevant to $W_a$, the problem is reduced to

\begin{equation}
\label{equ:obj2}
\begin{aligned}
\min_{W_a}
&
\left \{o_1(W_a)=
\sum_{t=1}^{T-1}\left(\sum_{i=1}^{n_t}  \| U_a^\top f_a(X_i^t) - \left ( \bfy_i^t - U_0^\top f_0(X_i^t) + U_t^\top f_t(X_i^t) \right ) \|_F^2\right)
\right .\\
&+\sum_{i=1}^{l_T}  \| U_a^\top f_a(X_i^T) - \left ( \bfy_i^T - U_0^\top f_0(X_i^T) + U_T^\top f_T(X_i^T) \right ) \|_F^2\\
&+ C_1 \sum_{t=1}^{T} \left ( \sum_{i=1}^{n_t} \left \|f_a(X_i^t) - U \bfa_i^t\right \|_F^2 \right )\\
&\left.
C_3 \sum_{i,i'=1}^{n_T} M_{ii'} \left ( \left \|f_a(X_i^T) - f_a(X_{i'}^T)\right \|_F^2
\right ) \right\}.
\end{aligned}
\end{equation}
To update the filters of attribute embedding function, we use the coordinate gradient descent algorithm. In this algorithm, the filters are updated sequentially. When one filter is updated, others are fixed. When the $k$-th filter $\bfw_{ak}$ (the $k$-th column of $W_a$) is considered, we update it according to the direction of the gradient of $o_1(W_a)$ regarding $\bfw_{ak}$. The gradient function is given as following according to chain rule,

\begin{equation}
\label{equ:gradient}
\begin{aligned}
&
\frac{\partial o_1}{\partial \bfw_{ak}}=
2\sum_{t=1}^{T-1}\sum_{i=1}^{n_t} \left[U_a \left ( U_a^\top f_a(X_i^t) - \left ( \bfy_i^t - U_0^\top f_0(X_i^t) + U_t^\top f_t(X_i^t) \right ) \right )  \right]_{k} \frac{\partial f_a(X_i^t)_k}{\partial \bfw_{ak}}  \\
&+2\sum_{i=1}^{l_T} U_a \left [ U_a^\top f_a(X_i^T) - \left ( \bfy_i^T - U_0^\top f_0(X_i^T) + U_T^\top f_T(X_i^T) \right ) \right ]_k \frac{\partial f_a(X_i^T)_k}{\partial \bfw_{ak}} \\
&+ 2C_1 \sum_{t=1}^{T} \sum_{i=1}^{n_t} \left [ f_a(X_i^t) - U \bfa_i^t \right ]_k \frac{\partial f_a(X_i^t)_k}{\partial \bfw_{ak}} \\
&+ 2C_3 \sum_{i,i'=1}^{n_T} M_{ii'} \left[ f_a(X_i^T) - f_a(X_{i'}^T)\right]_k \frac{\partial f_a(X_i^T)_k}{\partial \bfw_{ak}},
\end{aligned}
\end{equation}
where $[\bff]_k$ is the $k$-th element of the vector of $\bff$,

\begin{equation}
\label{equ:gradient_w}
\begin{aligned}
&\frac{\partial f_a(X)_k}{\partial \bfw_{ak}} =
\nabla g\left(\bfw_{ak}^\top \bfz_{j^*}\right) \bfz_{j^*},
\nabla g(x) =
\left\{\begin{matrix}
1, &if~x>0\\
0, &otherwise,
\end{matrix}\right.
~and\\
&
j^* = \underset{j}{\arg\max}(g(\bfw_{ak}^\top \bfz_i))),
\end{aligned}
\end{equation}
The updating rule of $\bfw_{ak}$ is as follows,

\begin{equation}
\label{equ:gradient1}
\begin{aligned}
\bfw_{ak}\leftarrow \bfw_{ak} - \tau \frac{\partial o_1}{\partial \bfw_{ak}},
\end{aligned}
\end{equation}
where $\tau$ is the descent step.

\subsubsection{Optimization of filters of $f_0(X)$}

To optimize the filters of the domain-independent convolutional representation function, $f_0(X)$, we also fix other parameters and remove the irrelevant terms. The following problem is obtained,

\begin{equation}
\label{equ:obj3}
\begin{aligned}
\min_{W_0}
&
\left \{o_2(W_0) =
\sum_{t=1}^{T-1}\sum_{i=1}^{n_t}
\| U_0^\top f_0(X_i^t) - \left ( \bfy_i^t - U_a^\top f_a(X_i^t) + U_t^\top f_t(X_i^t) \right ) \|_F^2\right.
\\
&\left.
+\sum_{i=1}^{l_T} \| U_0^\top f_0(X_i^T) - \left ( \bfy_i^T- U_a^\top f_a(X_i^T) + U_T^\top f_T(X_i^T) \right ) \|_F^2
\right .\\
&
+ C_2 \sum_{t,t'=1, t<t'}^{T} \left\| \frac{1}{n_t}\sum_{i=1}^{n_t} f_0(X_i^t)
- \frac{1}{n_{t'}}\sum_{i=1}^{n_{t'}} f_0(X_i^{t'})  \right \|_F^2\\
&\left.
+ C_3 \sum_{i,i'=1}^{n_T} M_{ii'} \left \|f_0(X_i^T) - f_0(X_{i'}^T)\right \|_F^2 \right\}.
\end{aligned}
\end{equation}
Similarly to the optimization of filters of $f_a(X)$, we also use the coordinate gradient descent algorithm to update the filters of $f_0(X)$. When a filter $\bfw_{0k}$ is considered, we calculate the sub-gradient function of $o_2(W_0)$ with regard to $\bfw_{0k}$ as follows,

\begin{equation}
\label{equ:gradient3}
\begin{aligned}
&
\frac{\partial o_2}{\partial \bfw_{0k}} = 2\sum_{t=1}^{T-1}\sum_{i=1}^{n_t}
\left[U_0\left( U_0^\top f_0(X_i^t) - \left ( \bfy_i^t - U_a^\top f_a(X_i^t) + U_t^\top f_t(X_i^t) \right ) \right)\right]_k \frac{\partial f_0(X_i^t)}{\partial \bfw_{0k}}
\\
&\left.
+2\sum_{i=1}^{l_T} \left[U_0\left(U_0^\top f_0(X_i^T) - \left ( \bfy_i^T- U_a^\top f_a(X_i^T) + U_T^\top f_T(X_i^T) \right ) \right)\right]_k \frac{\partial f_0(X_i^T)}{\partial \bfw_{0k}}
\right .\\
&
+ 2C_2 \sum_{t,t'=1, t<t'}^{T} \left[ \frac{1}{n_t}\sum_{i=1}^{n_t} f_0(X_i^t)
- \frac{1}{n_{t'}}\sum_{i=1}^{n_{t'}} f_0(X_i^{t'})  \right ]_k \\
&
\left(\frac{1}{n_t}\sum_{i=1}^{n_t}  \frac{\partial f_0(X_i^t)}{\partial \bfw_{0k}}
- \frac{1}{n_{t'}}\sum_{i=1}^{n_{t'}} \frac{\partial f_0(X_i^{t'})}{\partial \bfw_{0k}}  \right )
\\
&
+2 C_3 \sum_{i,i'=1}^{n_T} M_{ii'} \left [f_0(X_i^T) - f_0(X_{i'}^T)\right ]_k
\left ( \frac{\partial f_0(X_i^T)}{\partial \bfw_{0k}} - \frac{\partial f_0(X_{i'}^T)}{\partial \bfw_{0k}}  \right )
\end{aligned},
\end{equation}
where $\frac{\partial f_0(X)}{\partial \bfw_{0k}}$ is defined as the same as (\ref{equ:gradient_w}), and the update rule of $\bfw_{0k}$ is

\begin{equation}
\label{equ:update_w0}
\begin{aligned}
\bfw_{0k}\leftarrow \bfw_{0k} - \tau \frac{\partial o_2}{\partial \bfw_{0k}}.
\end{aligned}
\end{equation}

\subsubsection{Optimization of filters of $f_t(X), t=1, \cdots, T-1$}

To update the filters of a domain-specific convolutional representation function of an auxiliary domain, $f_t(X), t=1,\cdots, T-1$, we have the following optimization problem by fixing other parameters and removing the irrelevant terms,

\begin{equation}
\label{equ:obj4}
\begin{aligned}
\min_{W_t}
&
\left \{o_3(W_t) =
\sum_{i=1}^{n_t}  \|\left( U_t^\top f_t(X_i^t) - \left ( \bfy_i^t - U_a^\top f_a(X_i^t) +U_0^\top f_0(X_i^t)   \right ) \right)\|_F^2 \right\}.
\end{aligned}
\end{equation}
To optimize the filters, we also use the coordinate gradient descent algorithm. The gradient descent function of $o_3$ with regard to a filter of $W_t$, $\bfw_{tk}$ is given as follows,

\begin{equation}
\label{equ:gradient4}
\begin{aligned}
&
\frac{\partial o_3}{\partial \bfw_{tk}} = 2\sum_{i=1}^{n_t}  \left[U_t\left( U_t^\top f_t(X_i^t) - \left ( \bfy_i^t - U_a^\top f_a(X_i^t) +U_0^\top f_0(X_i^t)   \right ) \right)\right]_k
\frac{\partial f_t(X_{i}^t)}{\partial \bfw_{tk}},
\end{aligned}
\end{equation}
where $\frac{\partial f_t(X)}{\partial \bfw_{tk}}$ is defined as same as (\ref{equ:gradient_w}). Accordingly, $\bfw_{tk}$ is updated as

\begin{equation}
\label{equ:update_wt}
\begin{aligned}
\bfw_{tk}\leftarrow \bfw_{tk} - \tau \frac{\partial o_3}{\partial \bfw_{tk}}.
\end{aligned}
\end{equation}

\subsubsection{Optimization of filters of $f_T(X)$}

To update the filters of the target domain-specific convolutional representation function, $f_T(X)$, we only consider the terms of the objective function which are relevant to $f_T(X)$, and fix other parameters. The following optimization problem is obtained,

\begin{equation}
\label{equ:obj4}
\begin{aligned}
\min_{W_T}
&
\left \{o_4(W_T) =
\sum_{i=1}^{l_T}  \| U_T^\top f_T(X_i^T) - \left ( \bfy_i^T - U_a^\top f_a(X_i^T) +U_0^\top f_0(X_i^T) \right) \|_F^2
\right .\\
&\left.
+ C_3 \sum_{i,i'=1}^{n_T} M_{ii'} \left \|f_T(X_i^T) - f_T(X_{i'}^T)\right \|_F^2 \right\},
\end{aligned}
\end{equation}
and its gradient function with regard to the $k$-th filter is

\begin{equation}
\label{equ:gradient4}
\begin{aligned}
&
\frac{ \partial o_4}{\partial \bfw_{Tk}} =
2\sum_{i=1}^{l_T}  \left [ U_T \left( U_T^\top f_T(X_i^T) - \left ( \bfy_i^T - U_a^\top f_a(X_i^T) +U_0^\top f_0(X_i^T) \right) \right)\right]_k
\frac{ \partial f_T(X_i^T)}{\partial \bfw_{Tk}}
\\
&
+2 C_3 \sum_{i,i'=1}^{n_T} M_{ii'} \left [f_T(X_i^T) - f_T(X_{i'}^T)\right ]_k
\left ( \frac{ \partial f_T(X_i^T)}{\partial \bfw_{Tk}} - \frac{ \partial f_T(X_{i'}^T)}{\partial \bfw_{Tk}}\right).
\end{aligned}
\end{equation}
Accordingly, the update rule of $\bfw_{Tk}$ is

\begin{equation}
\label{equ:gradient5}
\begin{aligned}
\bfw_{Tk} \leftarrow \bfw_{Tk} - \tau \frac{\partial o_4}{\partial \bfw_{Tk}}.
\end{aligned}
\end{equation}

\subsubsection{Optimization of $U_a$, $U_t,t=0,\cdots, T$}

To optimize the transformation matrices $U_a$, $U_0$, and $U_t,t=1,\cdots, T$, we only consider the following optimization problem,

\begin{equation}
\label{equ:obj5}
\begin{aligned}
&
\min_{U_a,U_0,U_1,\cdots,U_T} \left \{ o_5(U_a,U_0,U_1,\cdots,U_T) =
\vphantom{\sum_1^2}
\right.\\
&\sum_{t=1}^{T-1}\sum_{i=1}^{n_t}  \|\bfy_i^t - \left( U_0^\top f_0(X_i^t) + U_t^\top f_t(X_i^t) + U_a^\top f_a(X_i^t)  \right) \|_F^2 \\
&\left .
+\sum_{i=1}^{l_T}  \|\bfy_i^T - \left( U_0^\top f_0(X_i^T) + U_t^\top f_t(X_i^T) + U_a^\top f_a(X_i^T) \right)\|_F^2
\right \}.
\end{aligned}
\end{equation}

\begin{itemize}
\item \textbf{Optimization of $U_a$} To optimize $U_a$, we rewrite the objective as follows,

\begin{equation}
\label{equ:obj_Ua}
\begin{aligned}
&
o_5(U_a,U_0,U_1,\cdots,U_T) =
\left \|
\Omega - U_a^\top F_a
\right \|_F^2,\\
&where~
F_a = [
\underset{n_1}{\underbrace{f_a(X_1^1), \cdots, f_a(X_{n_1}^1)}}
,\cdots,
\underset{l_T}{\underbrace{f_a(X_1^T), \cdots, f_a(X_{l_T}^T)}}
],\\
&\Omega = [
\underset{n_1}{\underbrace{\bfomega_1^1, \cdots, \bfomega_{n_1}^1}}
,\cdots,
\underset{l_T}{\underbrace{\bfomega_1^T, \cdots, \bfomega_{l_T}^T}}
],\\
&and~\bfomega_i^t =
\bfy_i^t - \left( U_0^\top f_0(X_i^t) + U_t^\top f_t(X_i^t)\right).
\end{aligned}
\end{equation}
We set the derivative of the object regarding to the $U_a$ to zero to obtain the optimal solution of $F_a$,

\begin{equation}
\label{equ:obj_Ua_derivative}
\begin{aligned}
&\frac{\partial o_5}{\partial U_a} =
-2 F_a\Omega^\top + 2 F_a {F_a}^\top U_a = 0\\
&\Rightarrow U_a = \left ( F_a F_a^\top\right )^{-1} F_a \Omega^\top.
\end{aligned}
\end{equation}

\item \textbf{Optimization of $U_0$}  To optimize $U_0$, we rewrite the objective as follows,

\begin{equation}
\label{equ:obj_U0}
\begin{aligned}
&
o_5(U_a,U_0,U_1,\cdots,U_T) =
\left \|
\Upsilon  - U_0^\top F_0
\right \|_F^2\\
&where ~
F_0 = [
\underset{n_1}{\underbrace{f_0(X_1^1), \cdots, f_0(X_{n_1}^1)}}
,\cdots,
\underset{l_T}{\underbrace{f_0(X_1^T), \cdots, f_0(X_{l_T}^T)}}
],\\
&\Upsilon = [
\underset{n_1}{\underbrace{\bfupsilon_1^1, \cdots, \bfupsilon_{n_1}^1}}
,\cdots,
\underset{l_T}{\underbrace{\bfupsilon_1^T, \cdots, \bfupsilon_{l_T}^T}}
],\\
&and~\bfupsilon_i^t =
\bfy_i^t - \left( U_a^\top f_a(X_i^t) + U_t^\top f_t(X_i^t)\right).
\end{aligned}
\end{equation}
We set the derivative regarding $U_0$ to zero and obtain the solution of $U_0$,

\begin{equation}
\label{equ:obj_U0_derivative}
\begin{aligned}
&\frac{\partial o_5}{\partial U_0} =
-2 F_0\Upsilon^\top + 2 F_0{F_0}^\top U_0 = 0\\
&\Rightarrow U_0 = \left ( F_0 F_0^\top\right )^{-1} F_0 \Upsilon^\top.
\end{aligned}
\end{equation}

\item \textbf{Optimization of $U_t|_{t=1}^T$}  To optimize $U_t$, we rewrite the objective as follows,

\begin{equation}
\label{equ:obj5_Ut}
\begin{aligned}
&
o_5(U_a,U_0,U_1,\cdots,U_T) \\
&=
\sum_{i=1}^{n_t}  \|\bfy_i^t - \left( U_0^\top f_0(X_i^t) + U_t^\top f_t(X_i^t) + U_a^\top f_a(X_i^t)  \right) \|_F^2 + R_t\\
&= \left \| \Pi_t - U_t^\top F_t\right \|_F^2 + R_t\\
&where~F_t = \left[f_t(X_1^t), \cdots, f_t(X_{n_t}^t) \right],\\
&\Pi_t = \left[
\bfpi_1^t, \cdots, \bfpi_{n_t}^t
\right],\\
&and~\bfpi_i^t = \bfy_i^t - \left( U_0^\top f_0(X_i^t) + U_a^\top f_a(X_i^t)  \right).
\end{aligned}
\end{equation}
where $R_t$ is combination of the terms which are irrelevant to $U_t$. By setting the derivative of the objective regarding $U_t$ to zero, we obtain the solution of $U_t$,

\begin{equation}
\label{equ:obj_Ut_derivative}
\begin{aligned}
&\frac{\partial o_5}{\partial U_t} =
-2 F_t\Upsilon^\top + 2 F_t{F_0}^\top U_t = 0\\
&\Rightarrow U_t = \left ( F_t F_t^\top\right )^{-1} F_t {\Pi_t}^\top.
\end{aligned}
\end{equation}

\end{itemize}

\subsubsection{Optimization of $\Theta$}

To optimize the attribute mapping matrix, $\Theta$, we fix the other parameters and consider the following sub-optimization problem.

\begin{equation}
\label{equ:obj_Theta}
\begin{aligned}
&\min_{\Theta} \left \{
o_6(\Theta)=
C_1 \sum_{t=1}^{T} \left ( \sum_{i=1}^{n_t} \left \|f_a(X_i^t) - \Theta^\top \bfa_i^t\right \|_F^2
\right ) \right.\\
&\left.
= C_1\left \| F_a -  \Theta^\top A \right \|_F^2
\right\},\\
&where~
A = [
\underset{n_1}{\underbrace{\bfa_1^1, \cdots, \bfa_{n_1}^1}}
,\cdots,
\underset{l_T}{\underbrace{\bfa_1^T, \cdots, \bfa_{l_T}^T}}
].
\end{aligned}
\end{equation}
By setting the derivative of $o_6$ regarding to $\Theta$ to zero, we obtain the optimal solution of $\Theta$ as follows,

\begin{equation}
\label{equ:obj_Theta_solution}
\begin{aligned}
&\frac{\partial o_6}{\partial \Theta}=
-2 C_1AF_a^\top + 2C_1 AA^\top \Theta= 0\\
&\Rightarrow \Theta = \left ( AA^\top\right)^{-1} A F_a^\top.
\end{aligned}
\end{equation}

\subsection{{Details of algorithm implementation}}

{In this section, we describe the details of the iterative algorithm for learning the parameters of the convolutional attribute embedding model, and the details of the implementation. The detailed description of the iterative algorithm is given in Algorithm \ref{alg:iter}. In this algorithm, we update the filters of three convolutional layers, the transformation matrices, and the attribute transformation matrix, alternately. At the very beginning of the algorithm, we initialize the parameters and an objective function value to zeros. The updating processes are iterated until a maximum iteration number or the amount of decreasing of the objective value is smaller than a threshold. The algorithm is implemented by Python programming language with Tensorflow supporting.}

\begin{algorithm}[h!]
\caption{Iterative algorithm of MRSO.}
\label{alg:iter}
\begin{algorithmic}
\STATE \textbf{Input}: Training set of data points $\{(X_i^t, \bfa_i^t, \bfy_i^t)|_{i=1}^{n_t}\}|_{t=1}^T$;
\STATE \textbf{Input}: Tradeoff parameters $C_1$, $C_2$, and $C_3$;
\STATE \textbf{Input}: Maximum number of iterations, $\eta$;
\STATE \textbf{Input}: Objective value threshold, $\varepsilon$.

\STATE Initialize iteration indicator $\iota = 1$.

\STATE Initialize model parameters and objective value $o^0=0$.

\WHILE{$\iota \leq \eta$ or objective value $|o^\iota - o^{\iota-1}|\leq \varepsilon$}

\STATE Update the filters of the convolutional attribute embedding model, $f_a$, according to (\ref{equ:gradient1}).

\STATE Update the filters of the domain-independent convolutional representation model, $f_0$, according to (\ref{equ:update_w0}).

\FOR{$i=t,\cdots,T-1$}

\STATE Update the filters of the domain-specific convolutional representation model of the $t$-th source domain, $f_t$, according to (\ref{equ:update_wt}).

\ENDFOR

\STATE Update the filters of the domain-specific convolutional representation model of the target domain, $f_T$, according to (\ref{equ:gradient5}).

\STATE Update the transformation matrix of attribute embedding model, $U_a$, according to (\ref{equ:obj_Ua_derivative}).

\STATE Update the transformation matrix of the domain-independent model,  $U_0$, according to (\ref{equ:obj_U0_derivative}).

\STATE Update the transformation matrix of the domain-specific model, $U_t|_{t=1}^T$, according to (\ref{equ:obj_Ut_derivative}).

\STATE Update the attribute mapping matrix, $\Theta$, according to (\ref{equ:obj_Theta_solution}).

\STATE Update the objective value $o^\iota$ according to (\ref{equ:obj1}).

\STATE $\iota = \iota+1$.

\ENDWHILE

\STATE \textbf{Output}:  $\bfw^T$ and $z_1^T,\cdots,z_n^T$.

\end{algorithmic}
\end{algorithm}

\section{Experiment}
\label{sec:exp}

In this section, we evaluate the proposed method over several domain-transfer problems.

\subsection{Data sets}

In the experiments, we use the following six data sets.

\begin{itemize}
\item \textbf{CUHK03 data set} This data set was developed for the problem of person re-identification problems \citep{li2014deepreid}. It contains 13,164 images of 1,360 persons. For each image, we annotate it by 108 attributes, including gender (male/female), wearing long hair, etc. The images are captured by six different cameras. The problem of person re-identification is to train a classifier over the images of some cameras, and then use the classifier to identify an image captured from other cameras. We treat each camera as a domain, and we use each domain as a target domain in turn. The dataset can be downloaded from \url{http://www.ee.cuhk.edu.hk/~xgwang/CUHK_identification.html}. {The attributes that we annotate in CUHK03 are listed in Table \ref{tab:Attributes}.}

{
\begin{table}
\centering
\caption{Attributes used to annotate CUHK03.}\label{tab:Attributes}
\begin{tabular}{|c|c|c|c|c|}
\hline
upperBodyRed & lowerBodyBrown & personalLess30 \\ \hline\hline
hairBrown & lowerBodyLogo & lowerBodyTrousers \\ \hline\hline
footwearShoes & carryingNothing & upperBodyBlue \\ \hline\hline
upperBodyBrown & upperBodyLogo & hairWhite \\ \hline\hline
hairRed & footwearPurple & personalLarger60 \\ \hline\hline
hairGrey & upperBodyWhite & lowerBodyHotPants \\ \hline\hline
carryingFolder & lowerBodyThinStripes & hairPurple \\ \hline\hline
upperBodyThinStripes & lowerBodyShorts & accessoryHeadphone \\ \hline\hline
footwearLeatherShoes & upperBodyPurple & footwearYellow \\ \hline\hline
upperBodyGrey & lowerBodyOrange & accessorySunglasses \\ \hline\hline
upperBodyLongSleeve & upperBodyOther & accessoryFaceMask \\ \hline\hline
accessoryMuffler & upperBodyNoSleeve & footwearBlue \\ \hline\hline
lowerBodyJeans & upperBodyOrange & upperBodyJacket \\ \hline\hline
hairGreen & footwearPink & lowerBodyShortSkirt \\ \hline\hline
personalLess45 & upperBodyFormal & carryingUmbrella \\ \hline\hline
footwearGreen & lowerBodyYellow & carryingBabyBuggy \\ \hline\hline
footwearRed & lowerBodyLongSkirt & hairYellow \\ \hline\hline
footwearBlack & lowerBodyWhite & lowerBodyGreen \\ \hline\hline
upperBodyYellow & footwearSandals & hairLong \\ \hline\hline
accessoryNothing & upperBodyThickStripes & upperBodyPlaid \\ \hline\hline
carryingPlasticBags & upperBodyShortSleeve & hairShort \\ \hline\hline
upperBodyCasual & accessoryKerchief & carryingSuitcase \\ \hline\hline
footwearSneakers & footwearGrey & upperBodyVNeck \\ \hline\hline
accessoryHat & hairOrange & personalLess60 \\ \hline\hline
accessoryHairBand & lowerBodySuits & upperBodySuit \\ \hline\hline
upperBodyBlack & footwearWhite & lowerBodyGrey \\ \hline\hline
carryingLuggageCase & lowerBodyCasual & upperBodyGreen \\ \hline\hline
carryingOther & lowerBodyPurple & footwearOrange \\ \hline\hline
upperBodyTshirt & lowerBodyRed & lowerBodyPlaid \\ \hline\hline
lowerBodyFormal & lowerBodyBlack & personalLess15 \\ \hline\hline
upperBodyPink & lowerBodyPink & personalMale \\ \hline\hline
hairBlack & carryingBackpack & footwearStocking \\ \hline\hline
footwearBrown & hairBald & personalFemale \\ \hline\hline
footwearBoots & accessoryShawl & lowerBodyBlue \\ \hline\hline
carryingShoppingTro & lowerBodyCapri & carryingMessengerBag \\ \hline
\end{tabular}
\end{table}
}

\item \textbf{Market-1501 data set} This data set is for the problem of person re-identification \cite{zheng2015scalable}. It contains 32,668 images of detected persons. The number of classes (identities) is 1,501. The number of cameras is six, and the number of cameras of images for each person varies from 2 to 6. The attribute set for this set is the same as the CUHK03 data set.

\item \textbf{iLIDS-VID data set} This data set is for the person re-identification problem \cite{wang2014person}. It contains images of 300 individuals, and for each individual, the image from two cameras are collected, thus there are 600 image sets in total. The number of images in each image set varies from 23 to 192.

\item \textbf{SAIVT-SoftBio data set} This data set of multiple camera person re-identification has images of 8 cameras \cite{bialkowski2012database}. It contains images of 152 different individuals in total. However, not all the individuals are detected in all the cameras, we only consider the persons who are detected in the 3-rd, 5-th, and 8-th cameras.

\item \textbf{Bankrupt prediction set} This data contains the stock price wave data of 3 years of 374 companies of three different countries, China, USA, and UK. We collected this data for the problem of prediction of company bankrupt. Each company is also labeled by a list of business type attributes. To represent the price change wave of a company, we use a sliding window to split the wave into short-term frames and treat each frame as an instance. In this way, each company is treated as a data points, presented by a set of short-term frames, and a list of binary attributes of business types. Moreover, each country is treated as domain, thus we have three domains in our setting. The prediction problem of this data set is to predict if a given company will be in bankrupt within the future 3 years. Again, we treat each country as a target domain in turn and use the other two countries as auxiliary domains.

\item \textbf{Spam email data set} This data set is for the spam email detection competition of the ECML/PKDD Discovery Challenge 2006 \citep{bickel2006ecml}. It contains texts of emails of 15 email users, and for each user, there are 400 emails. Among the 400 emails of each user, half of them are spam emails, while the remaining half are non-spam emails. Each email text is composed of a set of words. To present each email, we use the word embedding technology to obtain an embedding vector for each word of the email text, and thus each email is transformed to a set of embedding vectors, which is treated as instances in our model. Moreover, we also apply a topic classifier and a sentiment classifier to each email text to extract attributes of the text and use the extracted attributes as additional information. Each user is treated as a domain, and we also use each user as a target domain in turn.
\end{itemize}

{The motive to use the data sets of person re-identification, bankruptcy prediction, and spam email detection is to show that the proposed method can be generalized to different types of applications, including computer vision, natural language processing, and economics.}

\subsection{{Experimental setting}}

{
In our experiments, given a data set of several domains, we treat each domain as a target domain in turn, while treating the other domains as the auxiliary domains to help train the model. The data points in a target domain are further split into a training set and a test set with equal sizes randomly. Meanwhile, for the training set of the target domain, we further split it into equal-sized subsets. One subset is used as a labeled set, and the other set is used as an unlabeled set. We train the model over the data points of the auxiliary domains and the training set of the target domain and then test it over the test set of the target domain. The classification rate over the test set is used as the performance measure. The average classification rate over different target domains is reported and compared. The average classification rate is computed as follows,
}

{
\begin{equation}
\label{equ:acc}
\begin{aligned}
&average~classification~rate\\
&= \frac{1}{\#target~domains}\sum_{t}
\frac{\# correctely~ classified~data~ points~of~the~t-th~domain}{\#total~test~data~points~of~the~t-th~domain}
\end{aligned}
\end{equation}
}

\subsection{Results}

In the experiments, we first compare the proposed domain transfer convolutional attribute embedding (DTCAE) algorithm to some state-of-the-art domain-transfer attribute representation methods, and then study the properties of the proposed algorithm experimentally.

\subsubsection{Comparison to state-of-the-art}

\begin{figure}[!htb]
  \centering
  \includegraphics[width=0.7\textwidth]{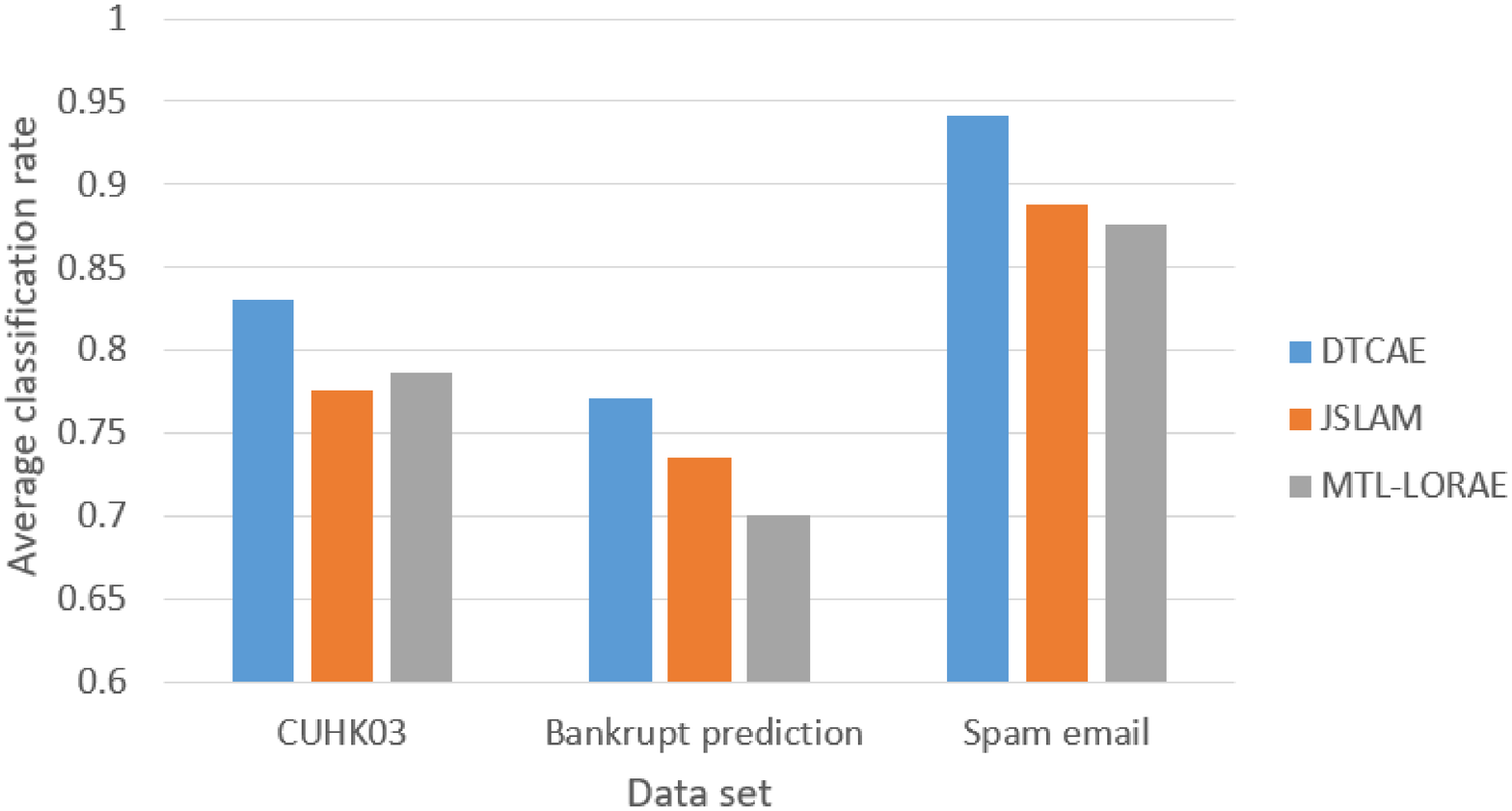}\\
  \includegraphics[width=0.7\textwidth]{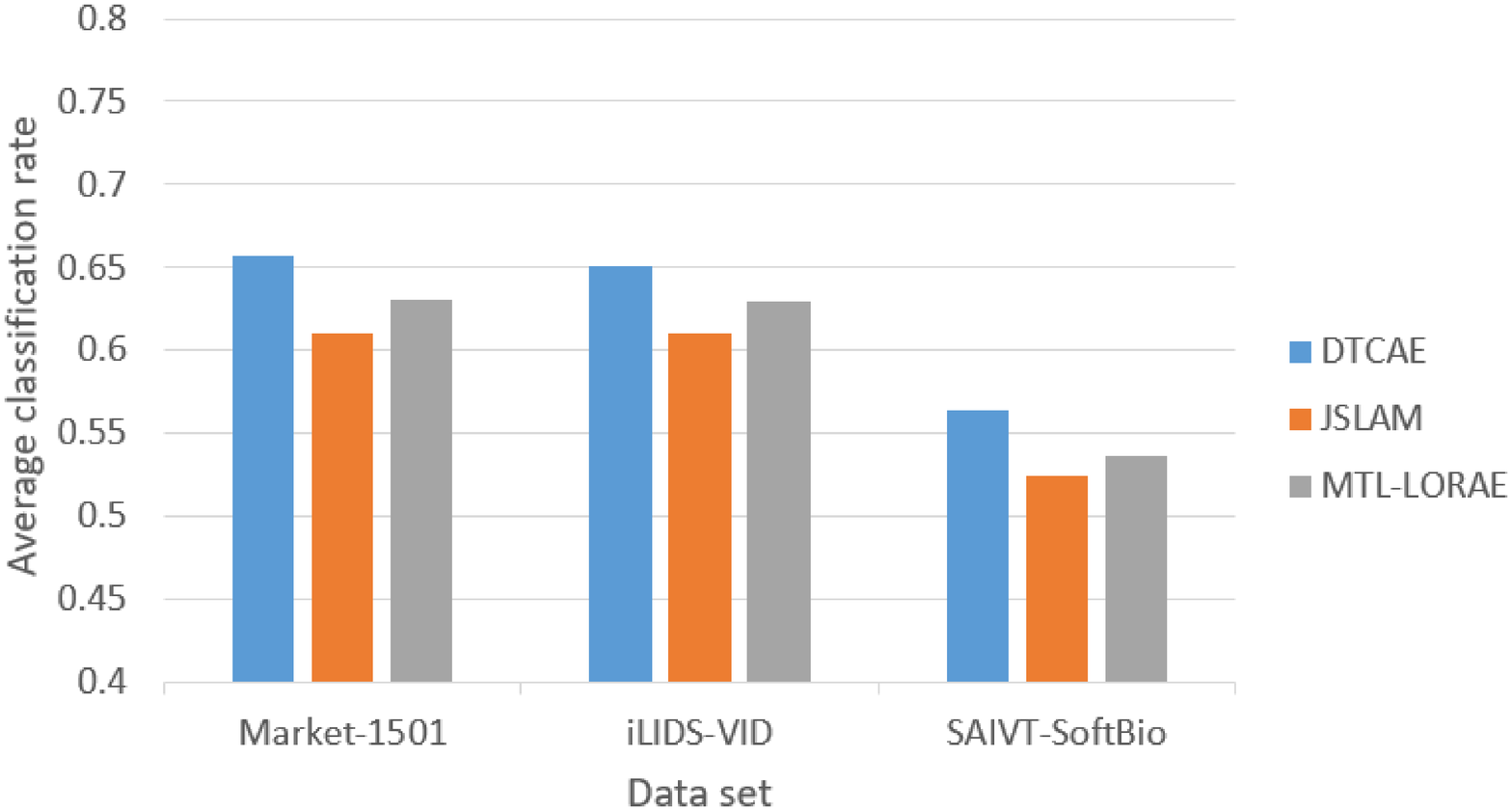}\\
  \caption{Comparison results over the benchmark data sets.}
  \label{fig:comparision}
\end{figure}

Attribute embedding for domain-transfer learning problem is a new topic and there are only two existing methods. In the experiment, we compare the proposed algorithm against the two existing methods, which are the Joint Semantic and Latent Attribute Modelling (JSLAM) method proposed by Peng et al. \citep{Peng2017}, and the Multi-Task Learning with Low Rank Attribute Embedding (MTL-LORAE) method proposed by Su et al. \citep{Su2017}. The comparison results over the three benchmark data sets are shown in Figure \ref{fig:comparision}. According to the reported average accuracies over the benchmark datasets, our algorithm DTCAE achieves the best performance over all the three datasets. For example, over the CUHK03 data set, the DTCAE is the only compared method which has an average accuracy higher than 0.800. Meanwhile, over the spam email dataset, only DTCAE obtains an average accuracy higher than 0.900. {The reasons for our improvement over the compare methods are described in detail as follows.}

\begin{itemize}
\item
{One reason for the improvement achieved by our model over the compared methods, JSLAM and MTL-LORAE, are the usage of convolutional attribute embedding layer. Both JSLAM and MTL-LORAE use simple linear functions to embed the attributes. These models heavily rely on the quality of the features extracted from the original data to represent the attributes. However, features are hand-crafted which is not specifically designed for the targeted attributes. However, our model is based on convolutional layers which use a group of filters to automatically extract features for the attributes recognition. The filters are adjusted to fit the attributes during the learning process. Compared to the linear model with hand-crafted features which ignores the attributes, the convolutional attribute embedding model can learn both the features and the attribute estimation function jointly. This makes the model more accurate for the attribute embedding.}

\item  {Another reason for the improvement is the usage of domain-independent and domain-specific convolutional representation layers to extract the features shared by different domains and the features specific for each domain. However, JSLAM ignores the difference between features of different domains and uses a shared dictionary to represent hand-crafted features extracted from the original features. Compared to JSLAM, our model has the advantage of automatic feature learning, and effective domain feature extraction and sharing. This makes our model more suitable for the domain-transfer learning problem.}

\end{itemize}

\subsubsection{Sensitivity to tradeoff parameters}

\begin{figure}[!htb]
\centering
\includegraphics[width=0.7\textwidth]{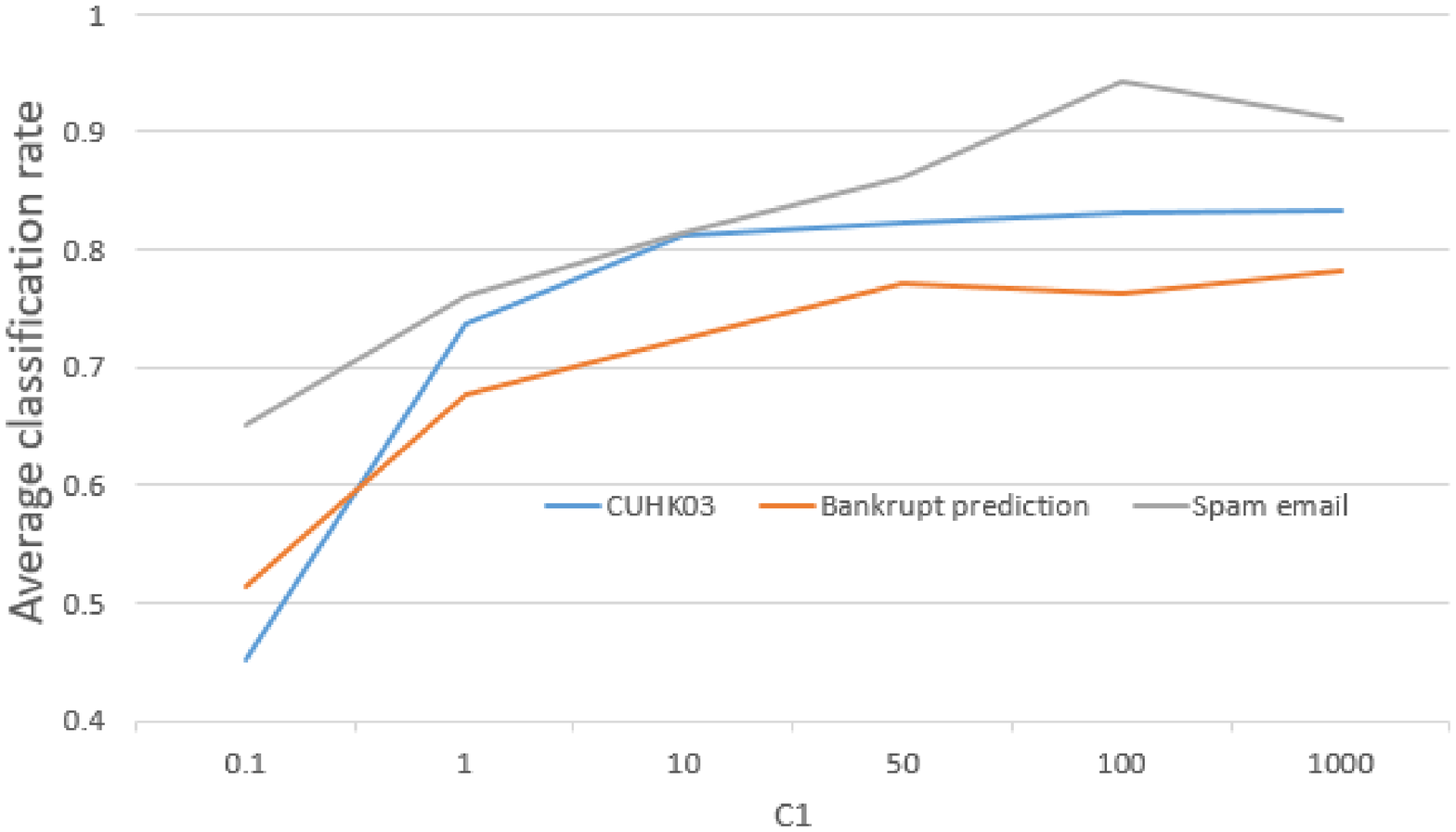}\\
\includegraphics[width=0.7\textwidth]{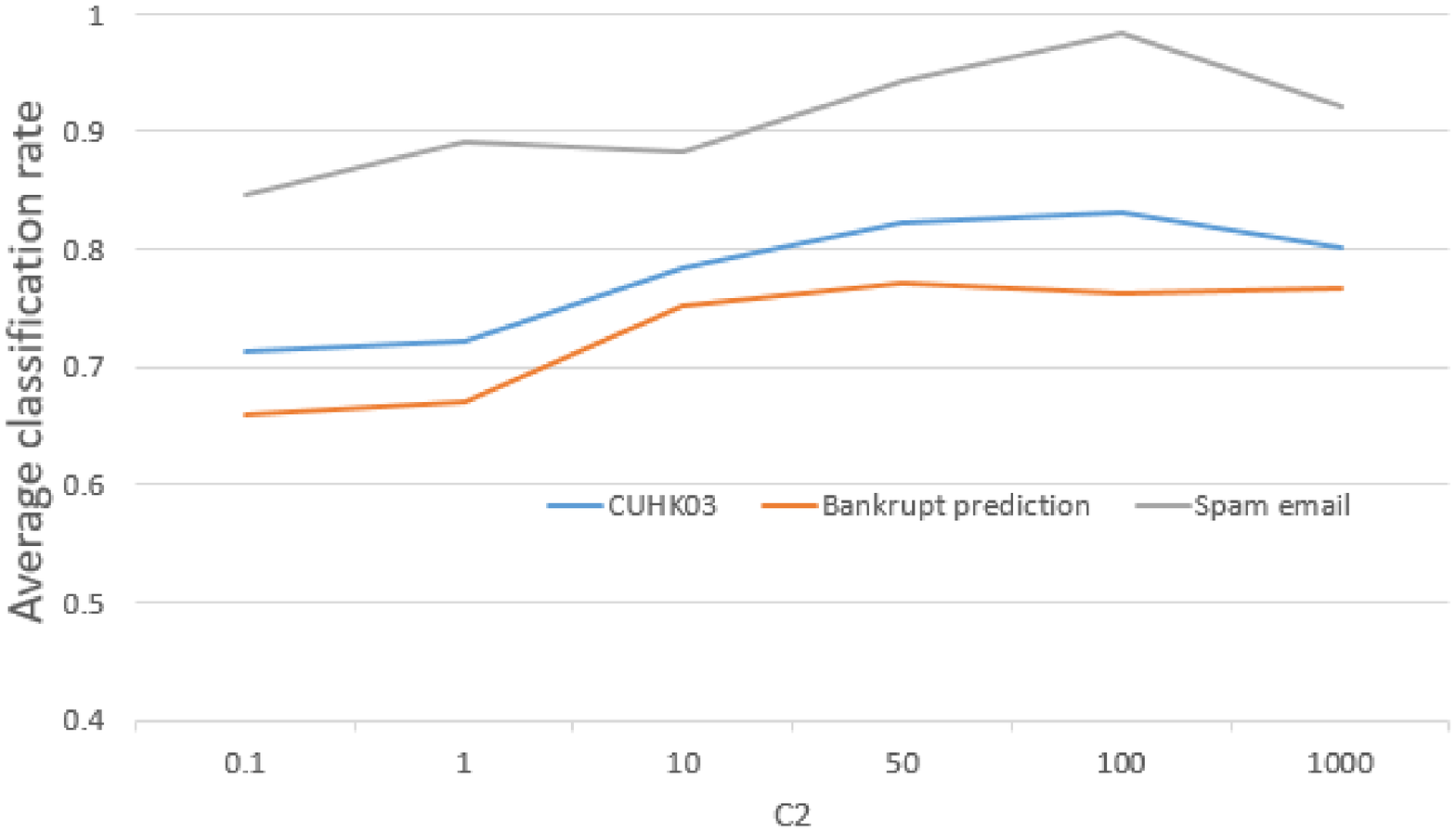}\\
\includegraphics[width=0.7\textwidth]{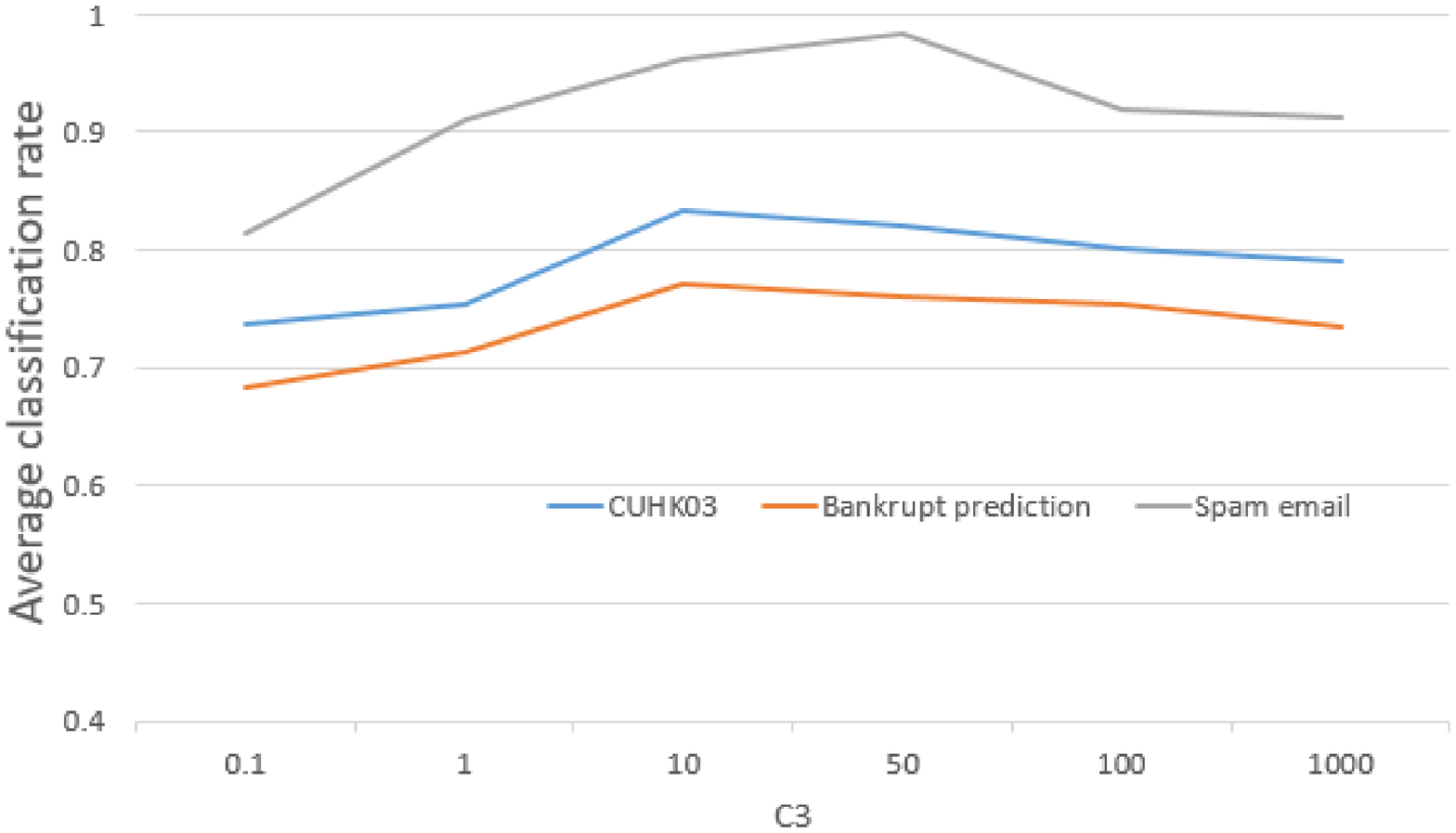}\\
  \caption{Sensitivity curve of tradeoff parameters.}
  \label{fig:parameter}
\end{figure}

In the objective function of our model, there are three tradeoff parameters, $C_1$, $C_2$, and $C_3$. These parameters weight the importance of attribute embedding, domain-independent representation, and neighborhood similarity regularization. To verify the effect of these terms, we also study the performance of the proposed algorithm against different values of the tradeoff parameters. The sensitivity curves of to the tradeoff parameters are reported in Figure \ref{fig:parameter}. As shown in the figure, for the $C_1$, our algorithm DTCAE is sensitive to the change of the value of $C_1$. When $C_1$ is increasing from 0.1 to 50, the average classification rates over all the three datasets increase significantly. This indicates the importance of the attributes embedding for the domain-transfer learning. However, it seems the proposed algorithm DTCAE is stable to the change of $C_2$. But a larger $C_2$ still achieves slightly better average classification rates. Finally, regarding $C_3$, we cannot observe a significant change when then values are varying. It seems a median value of $C_3$ can give the best results.

\subsubsection{Convergence analysis}

\begin{figure}[!htb]
  \centering
  \includegraphics[width=0.7\textwidth]{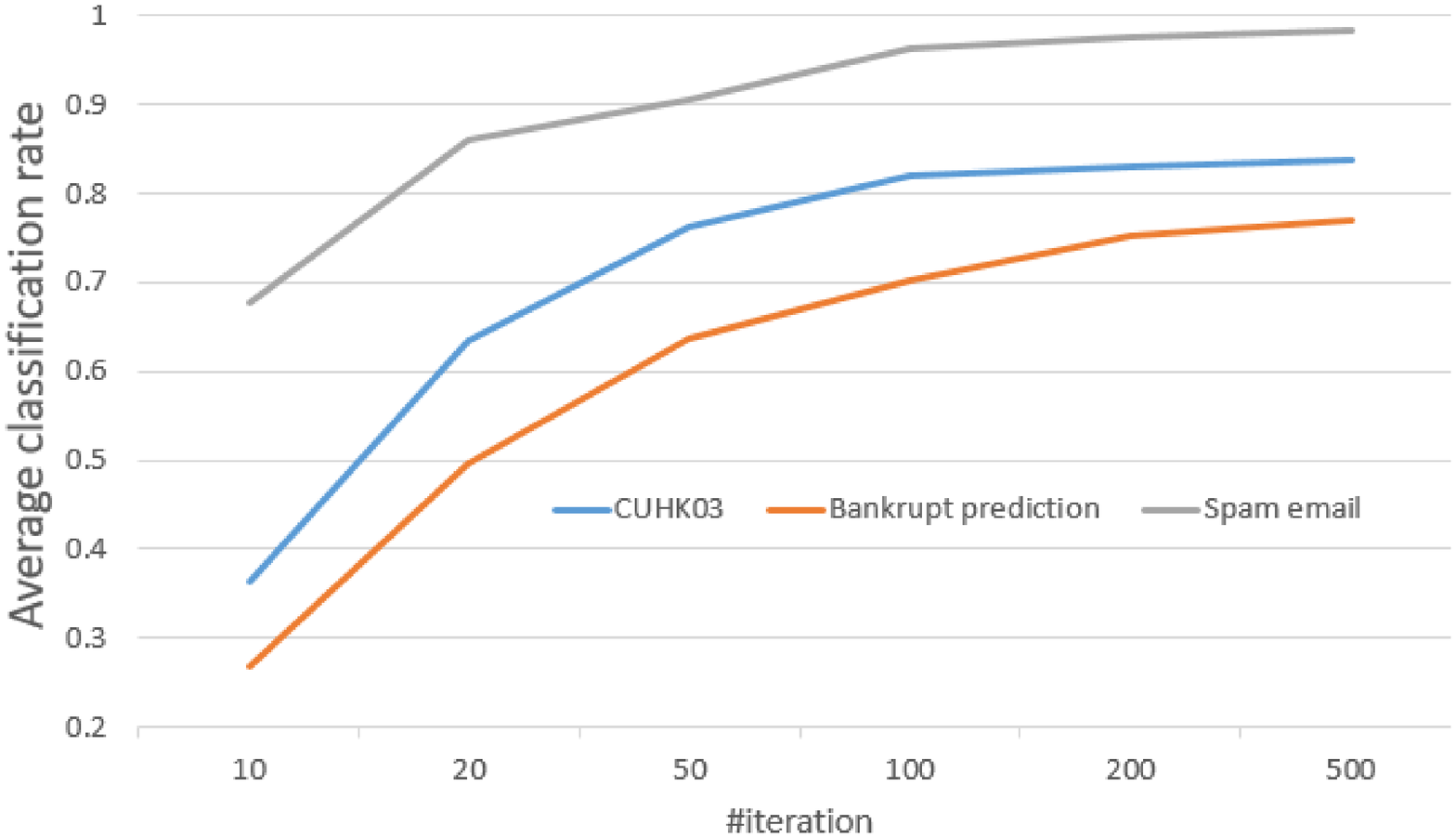}\\
  \caption{Convergence curves over the benchmark data sets.}
  \label{fig:curve}
\end{figure}

Since the proposed algorithm DTCAE is an iterative algorithm. The variables are updated alternately. We are also interested in the convergence of the algorithm. Thus we plot the average classification rates with a different number of iterations. The curves over the three benchmark data sets are plotted in Figure \ref{fig:curve}. According to the curves of Figure \ref{fig:curve}, when more iterations are used to update the variables of the model, the average classification rates increase stably. This is not surprising because a larger number of iterations reaches a smaller objective function. This verifies the effectiveness of the proposed model and its corresponding objective function. Moreover, we also observe that when the iteration number is larger than 100, the change of the performance is very small. This means that the algorithm converges and no more iteration is needed to improve the performance.

\section{Conclusion}
\label{sec:conclusion}

In this paper, we propose a novel model for the problem of cross-domain learning problem with attribute data. The model is based on CNN model. We use a CNN model to map the input data to its attributes. Moreover, a domain-independent and domain-specific CNN model are also used to represent the data input itself. The attribute embedding, the domain-independent, and domain-specific representations are concatenated as the new representation of the data points, and we further a linear layer to map the new representation to the class labels. Moreover, we also impose the domain-independent representations of data points of different domains to be in a common distribution, and the neighboring data points of target domain to be similar to each other. We model the learning problem as a minimization problem and solve it by an iterative algorithm. The experiments on three benchmark data sets show its advantages. In the future, we will consider apply the proposed method to more applications, such as Ad Hoc communications \cite{liu2012astra,cui2013leveraging,yang2017automatic,bhimani2018docker}, signal processing \cite{zhu2017sparse,zhu2016fast}, graph matching \cite{yu2016scene,yu2016graph}, biomedical imaging \cite{zhang2017empowering,zhang2016applying}, physics \cite{bai2010nonabelian,bai2010some,gasper2017adsorption}, medical informatics \cite{chen2017sequence,chen2017brief,}, etc.


\end{document}